\documentclass{article}

\PassOptionsToPackage{numbers, compress}{natbib}

\usepackage[preprint]{neurips_2022}
\usepackage{comment}

\usepackage[utf8]{inputenc} 
\usepackage[T1]{fontenc}    
\usepackage{hyperref}       
\usepackage{url}            
\usepackage{booktabs}       
\usepackage{amsfonts}       
\usepackage{nicefrac}       
\usepackage{microtype}      
\usepackage{xcolor}         
\usepackage{graphicx}
\usepackage{amsmath}
\usepackage{subcaption}
\usepackage{color}
\PassOptionsToPackage{usenames,dvipsnames}{color}
\usepackage{caption} 
\usepackage[toc,page]{appendix}
\usepackage[hang]{footmisc}

\title{Large language models converge toward human-like concept organization}

\author{
  \and 
  \small{\textbf{Jonathan Gabel Christiansen}\thanks{Equal contributions} $\;$ and \textbf{Mathias Lykke Gammelgaard$^*$} and \textbf{Anders Søgaard}} \\
  \small{Department of Computer Science}\\
  \small{University of Copenhagen}\\
  \small{Corresponding author: \texttt{soegaard@di.ku.dk}}
}

\begin{document}

\maketitle

\begin{abstract}
 Large language models show human-like performance in knowledge extraction, reasoning and dialogue, but it remains controversial whether this performance is best explained by memorization and pattern matching, or whether it reflects human-like inferential semantics and world knowledge. Knowledge bases such as WikiData provide large-scale, high-quality representations of inferential semantics and world knowledge. We show that large language models learn to organize concepts in ways that are strikingly similar to how concepts are organized in such knowledge bases. Knowledge bases model collective, institutional knowledge, and large language models seem to induce such knowledge from raw text. We show that bigger and better models exhibit more human-like concept organization, across four families of language models and three knowledge graph embeddings. 
\end{abstract}

\section{Introduction}
The artificial intelligence community is split on the question of whether

\begin{quote}
    “some generative
model [i.e., language model] trained only on text, given enough data and computational resources,
could understand natural language in some non-trivial sense.”
\end{quote}

Half of the community (51\%) -- according to a recent survey \citep{doi:10.1073/pnas.2215907120} -- are willing to attribute non-trivial understanding to large language models (LLMs). The other half of the community (49\%) argue that the illusion of understanding is the result of an Eliza effect.\footnote{One famous example of this view is an article by Emily Bender and colleagues \cite{bender2021dangers}, who argue that these models are simply \textit{stochastic parrots} that `haphazardly stitch together sequences of linguistic forms' without any true understanding of the world or context.} The research question, as formulated by \citet{doi:10.1073/pnas.2215907120}, is: 

\begin{quote} "do these systems (or will their near-term successors) actually, even in the absence
of physical experience, create something like the rich concept-based mental models that are central
to human understanding, and, if so, does scaling these models create even better concepts?"
\end{quote}

We present a series of experiments designed to answer this question directly. Our findings suggest (very strongly) that the models (representations) induced by larger and better LLMs become more and more human-like. 

\begin{figure}[h]
    \centering
    \includegraphics[width=\textwidth]{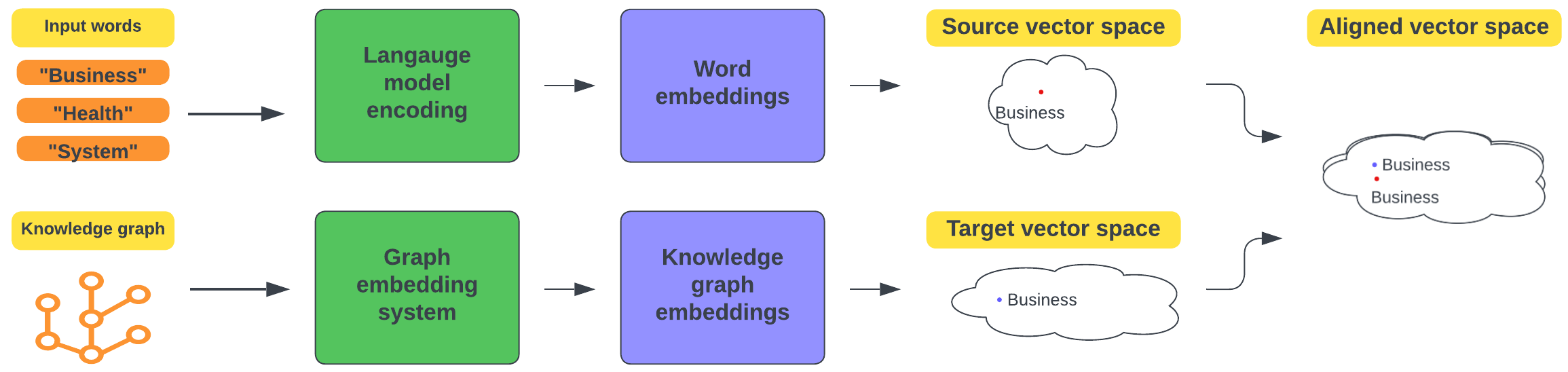}
    \caption{A simplified sketch of our experimental protocol. A vocabulary of 20K words is encoded using a language model and the corresponding entities are fetched from a pre-trained graph embedding system. The resulting vector spaces are then aligned. After alignment we evaluate retrieval performance in the target vector space. If retrieval performance is perfect, the spaces are (nearest neighbor graph) isomorphic.}
    \label{fig:my_label}
\end{figure}

\subsection{Contributions}
We present a series of experiments with four families of LLMs (21 models), as well as three knowledge graph embedding algorithms. Using three different methods, we compare the vector spaces of the LLMs to the vector spaces induced by the graph embedding algorithms. (This amounts to a total of 220 experiments.) We find that the vector spaces of LLMs within each family become increasingly structurally similar to those of knowledge graph embeddings. This shows that LLMs partially converge on human-like concept organization, facilitating inferential semantics \cite{piantadosi2022meaning}.\footnote{We use `converge' in the sense of \citet{caucheteux2022}.} The sample efficiency of this convergence seem to depend on a number of factors, including polysemy and semantic category. Our findings have important implications. They vindicate the conjecture in \cite{piantadosi2022meaning} that LLMs exhibit inferential semantics, settling the research question presented in \cite{doi:10.1073/pnas.2215907120}, cited above. This means that LLMs partially converge toward human-like concept representations and, thus, come to partially `understand language', in a non-trivial sense. We speculate that the human-like conceptual organization is also what facilitates out-of-distribution inferences in LLMs. 

\section{Experiments}
\subsection{Language models}
We evaluate the vector spaces induced by four well-known families of language models, conducting experiments with a  total of 20 different transformer-based models, as well as a baseline static word vector space. The four families are OPT \citep{zhang2022opt}, GPT-2 \citep{noauthororeditor}, and Pythia \citep{biderman2023pythia} (non-deduplicated version at model checkpoint step 143000) and BERT \cite{devlin2019bert}.\footnote{We also evaluated the T5 \cite{2020t5} LM series. T5 is trained with a multi-task objective, leading to mixed results.} We also evaluated the vector space of GPT-3 \citep{brown2020language}, i.e.,  \textit{text-embedding-ada-002}.\footnote{\href{https://platform.openai.com/docs/guides/embeddings/what-are-embeddings}{{https://platform.openai.com/docs/guides/embeddings/what-are-embeddings}}}
Transformer-based LLMs use multiple layers of self-attention \citep{vaswani2017attention} and can model complex interactions across large context windows. Both left and right context can be considered. GPT, OPT and Pythia are decoder-only autoregressive LLMs, however, and thus only consider left context, i.e., the words preceding the next token. BERT is an encoder-only non-autoregressive LLM and considers both left and right context of the masked token to be predicted. Transformers are in general considered state-of-the-art for most \texttt{NLP} tasks \citep{wolf-etal-2020-transformers}.
For each of the language model families, we consider variants with increasing size in terms of the number of model parameters. See Table \ref{tab:LM-stats} for a model overview.
\begin{table}[h]
\centering
\begin{tabular}{@{}llllllll@{}}
\toprule
\textbf{LM} & \textbf{Params}                   & \textbf{LM}           & \textbf{Params}           & \textbf{LM} & \textbf{Params}  & \textbf{LM} & \textbf{Params}\\ \midrule
OPT-\tiny{125M}     & \multicolumn{1}{l}{125M} & GPT-2 \tiny{small}   & 117M    & Pythia-\tiny{70M}   & \multicolumn{1}{l}{70M}   & BERT-\tiny{TINY}   & \multicolumn{1}{l}{4.4M}  \\
OPT-\tiny{350M}     & \multicolumn{1}{l}{350M} & GPT-2 \tiny{medium}  & 345M    & Pythia-\tiny{160M}  & \multicolumn{1}{l}{160M}  & BERT-\tiny{MINI}   & \multicolumn{1}{l}{11.3M}  \\
OPT-\tiny{1.3B}     & \multicolumn{1}{l}{1.3B} & GPT-2 \tiny{large}   & 774M    & Pythia-\tiny{410M}  & \multicolumn{1}{l}{410M}  & BERT-\tiny{SMALL}  & \multicolumn{1}{l}{29.1M}  \\
OPT-\tiny{2.7B}     & \multicolumn{1}{l}{2.7B} & GPT-2 \tiny{xl}      & 1.5B   & Pythia-\tiny{1B}    & \multicolumn{1}{l}{1B}    & BERT-\tiny{MEDIUM} & \multicolumn{1}{l}{41.7M}  \\
OPT-\tiny{6.7B}     & \small{6.7B}             & GPT-3 \tiny{Ada-002} & 175B    & Pythia-\tiny{2.8B}  & \multicolumn{1}{l}{2.8B}  & BERT-\tiny{BASE}   & \multicolumn{1}{l}{110.1M}  \\
                     &                          &                       &         & Pythia-\tiny{6.9B}  & \multicolumn{1}{l}{6.9B}  &                     &                          \\\bottomrule
\end{tabular}
\caption{21 transformer-based language models used in this experiment. 
}
\label{tab:LM-stats}
\end{table}

\subsection{Knowledge graph embeddings}

We experiment with three graph embedding algorithms and the vector spaces induced by running these on large-scale knowledge bases.
\subsubsection{BigGraph}
\label{sec:biggraph}
 The first vector space 
is that of the so-called \emph{BigGraph} embeddings \citep{https://doi.org/10.48550/arxiv.1903.12287}. BigGraph is trained on an input knowledge graph, i.e. a list of edges, identified by its' source and target entities and a relation type. The network output is a feature vector or embedding for every entity in the graph. An inherent quality of this method is that adjacent entities are placed close to each other in the vector space. The particular embeddings used in this work are obtained by pre-training on WikiData, a well-known knowledge base\footnote{\href{https://torchbiggraph.readthedocs.io/en/latest/pretrained_embeddings.html}{\tiny{https://torchbiggraph.readthedocs.io/en/latest/pretrained\_embeddings.html}}}. Knowledge bases like Wikidata provide a structured representation of the real-world \citep{knowledgebase} and encode implicit world knowledge.
The BigGraph embeddings contain all entities from the \emph{"truthy"} Wikidata dump (2019-03-06) and thus includes URLs, dates etc; which are not (directly) included in language model vocabularies. To ensure compatability between the vector spaces, we limit ourselves to \textit{single word} BigGraph entities. From these single word entities we pick $20,000$ common English words.\footnote{\href{https://github.com/first20hours/google-10000-english/blob/master/20k.txt}{\tiny{https://github.com/first20hours/google-10000-english/blob/master/20k.txt}}}

\subsubsection{GraphVite}

Our second and third knowledge base-derived vector spaces were both obtained by the GraphVite graph embedding algorithm \citep{Zhu_2019}. We use the following pre-trained models; \textit{TransE} \citep{conf/nips/BordesUGWY13} and  \textit{ComplEx} \citep{trouillon2016complex}, both of which are pre-trained on WikiData5m \citep{wang2020kepler}. We use the same entities presented in §\ref{sec:biggraph}.
\subsection{Graph isomophism}
An isomorphism from $G_1 = (V_1, E_1)$ to $G_2 = (V_2, E_2)$ is a bijection $f:V_1\rightarrow V_2$ such that any pair of nodes $a$ and $b$ are joined by an edge iff $f(a)$ and $f(b)$ are joined by an edge. 
Near-isomorphism of graphs refers to the situation where two graphs are not exactly isomorphic but exhibit strong structural similarity. Note that if the nearest neighbor graphs of two embedding spaces are isomorphic, there exists a vector space mapping with precision@1 of 1.0; see §2.7 for details on how to compute precision@1. We will evaluate to what extent (the $k$-nearest neighbor graphs of the) LLM vector spaces are isomorphic to the knowledge graph embeddings, by computing representational similarity analysis scores, as well as by evaluating the precision@$k$ of linear projections.  

\subsection{Linear projections}
We present two distinct methods of mapping (or projecting) the vector space of an LLM to the vector space of BigGraph (BG) and GraphVite (GV). Let $\mathbf{M}_{\texttt{LM}} \in \mathbf{R}^{V \times d_{\text{e}}}$ be a matrix of word embeddings and $\mathbf{M}_{\texttt{REF}} \in \mathbf{R}^{V \times d_{\texttt{ref}}}$ a matrix of knowledge graph node embeddings, where $V$ denotes the size of the vocabulary, $d_{\text{e}}$ the dimensionality of the word embeddings for a given language model and $d_{\text{ref}}$ refers to dimensionality of the knowledge graph embeddings, with $d_{\text{ref}} = 200$ for BG and $d_{\text{ref}} = 512$ for GV.

Our first approach is to use generalized Procrustes analysis \citep{RePEc:spr:psycho:v:31:y:1966:i:1:p:1-10} to align $\mathbf{M}_{\texttt{LM}}$ with $\mathbf{M}_{\texttt{REF}}$. Since this method enforces $d_e = d_{\text{ref}}$, we use \texttt{PCA} to reduce the dimensionality of $\mathbf{M}_{\texttt{LM}}$ to the desired size. The aim of Procrustes analysis is thus to find a transformation matrix $\mathbf{\Omega}$ that minimizes the sum of squared distance between each pair of word embeddings in $\mathbf{M}_{\texttt{LM}}$ with $\mathbf{M}_{\texttt{REF}}$. This is achieved by solving the following problem:
 \begin{align*} 
&\min_{\mathbf{\Omega} = s\mathbf{A}} ||\mathbf{\Omega}\mathbf{M}_{\texttt{LM}} - \mathbf{M}_{\texttt{REF}}||_F^2\\
&s \in \mathbf{R}^+,\; \mathbf{A} \in \mathbf{R}^{d_e\times d_e}\; \text{s.t.}\; \mathbf{A^TA} = \mathbf{I}
\end{align*}
Where $F$ denotes the Frobenius norm and we have that $\mathbf{\Omega}$ can be computed using singular value decomposition. In practice, we compute $\mathbf{\Omega}$ using a subset of the full vocabulary; $V_{\texttt{train}}$.

Secondly, we propose utilizing $d_{\text{ref}}$ ridge regression models $f_i$ for $i = 1,2,..,d_{\text{ref}}$, with one predictor for each dimension of the reference vector space. Each predictor $f_i$ is trained on a subset of the full vocabulary $V_{\texttt{train}}$ and learns a function $f_i : \mathbf{R}^{d_e} \rightarrow \{\mathbf{R}\}_j$, for $j = 1,2,..,d_{\text{ref}}$, where $j$ indicates the \textit{j}'th dimension of the reference vector space. The ridge regression models are re-trained for each language model as $d_e$ varies across these. Once trained, the models can be used to project the remaining vocabulary $V_{\texttt{test}}$. The methodology of using a separate ridge regression model for each dimension of the reference/target vector space has previously been used to decode linguistic meaning from brain activation \citep{ridge}.

\subsection{Representational Similarity Analysis} To further gauge the similarity of the vector spaces induced by LLMs and knowledge graph embeddings, we present experiments using Representational Similarity Analysis (RSA) \citep{rsa}. For a given language model we consider the matrix $\mathbf{M}_{\texttt{LM}}$ of word embeddings alongside the corresponding matrix $\mathbf{M}_{\texttt{REF}}$. We compute the representational dissimilarity matrices (RDMs); i.e. for each word embedding in each of the matrices we compute the euclidean distance to all other word embeddings within that respective matrix, thus generating two $V \times V$ RDMs. Once the RDMs has been computed (denote them $r_1$ and $r_2$), they are then compared used cosine similarity:
\[\text{cos}(r_1,r_2) = \frac{r_1^Tr_2}{\sqrt{r_1^Tr_1r_2^2r_2}}\]
A cosine similarity close to 1.0 will indicate $\mathbf{M}_{\texttt{LM}}$  more closely resembles $\mathbf{M}_{\texttt{REF}}$. Note that we flatten the RDMs in practice, to get a single value as a final metric. Representational similarity analysis is a well-known method within neuroscience, see for instance \citep{rsa1}.

\subsection{Analogies}

Finally, we carry out experiments using the WiQueen analogy dataset \citep{Garneau2021AnalogyTM}. The dataset consists of quadruples <$w_1,w_2,w_3,w_4$> of analogies, e.g. <\textit{Hefei,Anhui,Guiyang,Guizhou}> which corresponds to the analogy \textit{"\textbf{Hefei} is to \textbf{Anhui}, as \textbf{Guiyang} is to \textbf{Guizhou.}"}. We encode the individual words from the analogies using each of the language models, thus obtaining a new quadruple <$e_1,e_2,e_3,e_4$> of word embeddings for each analogy and language model. We then proceed to \textit{"solve"} the analogy mathematically by computing $e_1 - e_2 + e_3 = e_{\text{new}}$ 
\citep{ethayarajh2019rotate}. Finally, we compare  $e_{\text{new}}$ to $e_4$, by checking if $e_{\text{new}}$ and  $e_4$ are nearest neighbors in the WiQueen vector space. Note how for an LLM vector space to solve the analogy task (completely) is equivalent to being (nearest neighbor graph) isomorphic to the underlying knowledge base \cite{peng2022understanding}. 

\subsection{Evaluation}
For the experiments in which we induce a linear mapping $\mathbf{\Omega}$ and $f(\mathbf{M}_{\texttt{LM}})$ between the vector spaces of LLMs and  BigGraph/GraphVite, we evaluate how close $\mathbf{M}_{\texttt{LM}}\mathbf{\Omega}$ and $f(\mathbf{M}_{\texttt{LM}})$ is to $\mathbf{M}_{\texttt{REF}}$ using precision@\textit{k} as our performance metric.\footnote{This is the {\em de facto}~standard performance metric in the word vector space alignment literature \cite{4264a46fd9e846e4a704b2d13002e521}.} That is, for each word $w$ in $V_{\text{test}}$, we perform k-nearest neighbors of the corresponding word embedding contained within the projection of $\mathbf{M}_{\texttt{LM}}$ in the reference vector space. If $w$ is found among the k-nearest neighbors in the reference vector space, we say that the precision at $k$ (p@k) is 100\%. The final precision is then scored as an average over all words in $V_{\text{test}}$. Note that for all values of $k$ this will either be a \textit{"hit"} or a \textit{"miss"} as there is only one relevant item to retrieve. The full vocabulary $V$ contains 20,000 words and is split into $V_{\text{train}}$ and $V_{\text{test}}$ at 80\%/20\% respectively, which makes a random retrieval baseline $\texttt{P@1} = \frac{1}{4000}$. In practice,  our linear projections are found to be significantly more precise, which in turn reflects the growing resemblance between the vector spaces of increasingly larger language models and the vector spaces of the knowledge graph embeddings induced by the BigGraph/GraphVite graph embedding algorithms. Note that for the experiment involving analogies, we do not use a reference vector space, but instead evaluate the retrieval directly using the WiQueen data set.

For the representational similarity analysis, we simply use the cosine similarity between the RDM of each language model and the RDM of BigGraph/GraphVite as the performance metric.
\section{Results}

\subsection{Procrustes Analysis}
\label{sec:procrustes}

\begin{figure}[h]
    \centering
    \includegraphics[width=13cm]{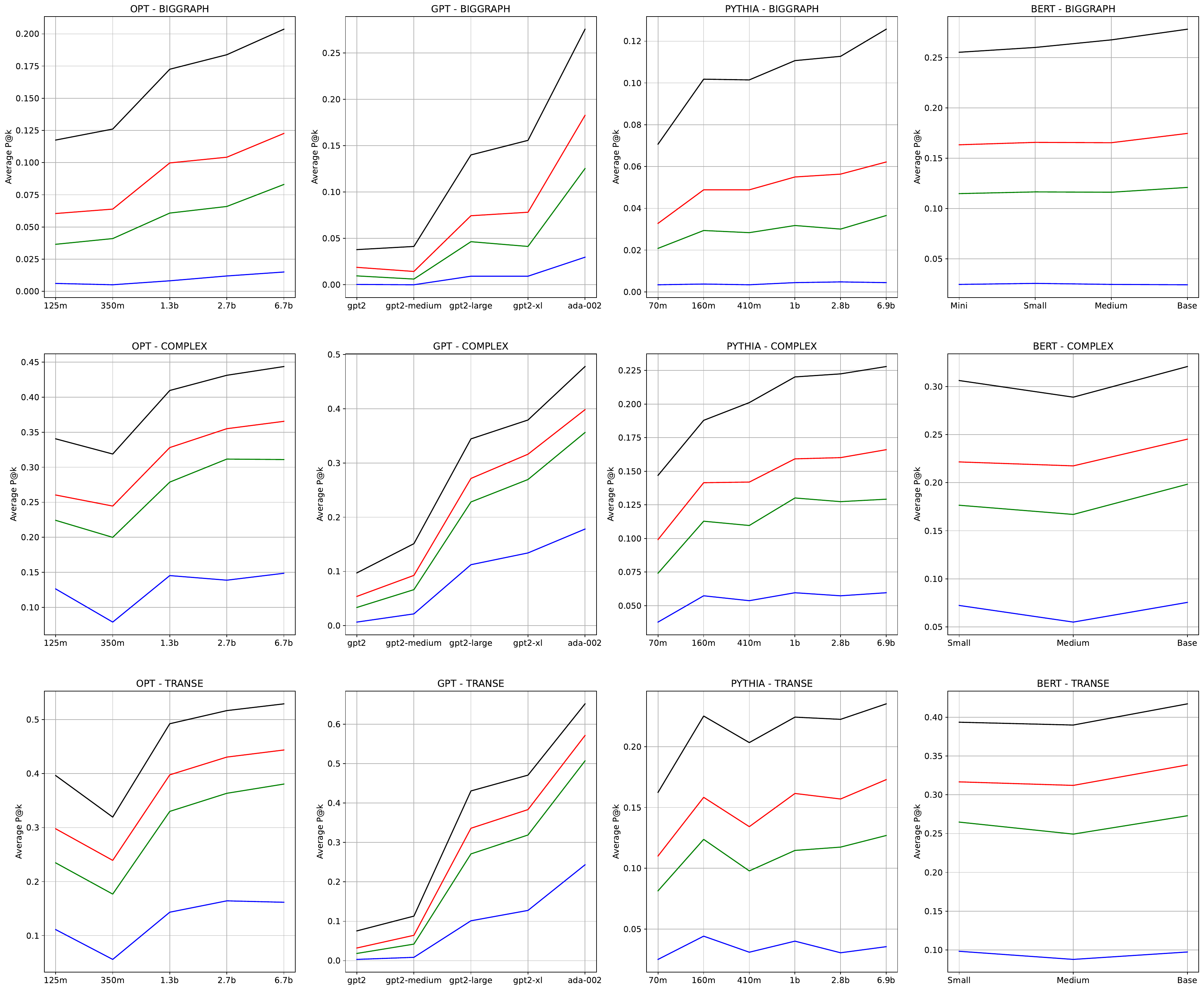}
    \caption{Plot labels: \textcolor{blue}{\rule[0.08cm]{0.5cm}{1.0pt} $k=1$}, \textcolor{green}{\rule[0.08cm]{0.5cm}{1.0pt} $k=10$}, \textcolor{red}{\rule[0.08cm]{0.5cm}{1.0pt} $k=20$}, \textcolor{black}{\rule[0.08cm]{0.5cm}{1.0pt} $k=50$}. Projection onto the vector space of BigGraph (1st row), ComplEx (2nd row) and TransE (3rd row) using \textbf{generalized Procrustes analysis} and retrieval performance p@k at $k=\{1,10,20,50\}$ for 4 language model families. We see results up to p@50$\sim$0.7, and strong, positive convergence (almost) across the board.}
    \label{procrustes}
\end{figure}

We report alignment precision (p@k) for $k \in \{1,10,20,50\}$, with our main results depicted in Figure~\ref{procrustes}. The plots show the convergence results, i.e. the relationship between language model size and alignment precision. For all four model families, we see a consistent trend, where larger language models lead to better alignments with the reference vector spaces. Overall, GPT appears to have the most pronounced convergence properties. Note that the different graph embeddings heavily influence the precision, but that the \textit{trend} is similar across graph embedding systems. For our best performing model GPT-3 (ada-002) projected onto the vector space of GraphVite (TransE) using Procrustes analysis, we observe a P@50 beyond 60\%, which in turn means that more than 6/10 words are mapped to a relatively small neighborhood of 50 words out of a total of 4000 words in $V_{\text{test}}$. This, in our view, constitutes strong evidence that LLMs learn human-like concept organizations.

\subsection{Ridge Regression}
\label{sec:ref}

\begin{figure}[h]
    \centering
    \includegraphics[width=13cm]{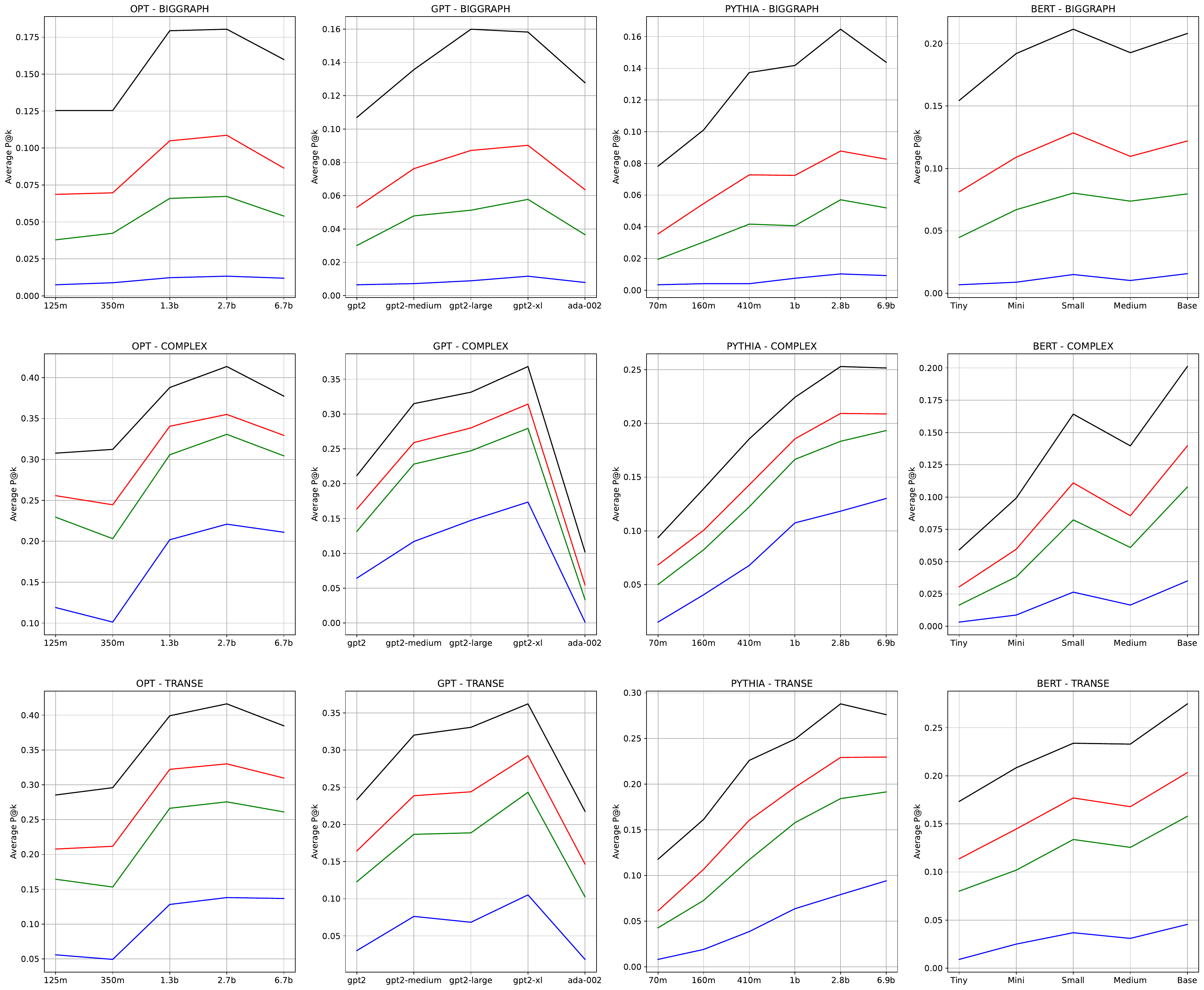}
    \caption{Plot labels: \textcolor{blue}{\rule[0.08cm]{0.5cm}{1.0pt} $k=1$}, \textcolor{green}{\rule[0.08cm]{0.5cm}{1.0pt} $k=10$}, \textcolor{red}{\rule[0.08cm]{0.5cm}{1.0pt} $k=20$}, \textcolor{black}{\rule[0.08cm]{0.5cm}{1.0pt} $k=50$}. Projection onto the vector space of BigGraph (1st row), ComplEx (2nd row) and TransE (3rd row) using \textbf{ridge regression} and retrieval performance p@k at $k=\{1,10,20,50\}$ for 4 language model families. We see strong, positive convergence, except for the GPT-3 results.
    }
    \label{fig:ridge}
\end{figure}

As a secondary projection method we train $d_{\text{ref}}$ ridge regression predictors, i.e. one for each dimension of the reference vector space. These predictors are then used to project the remaining vocabulary of word embeddings $V_{\text{test}}$ for a given language model to the reference vector space. After this, retrieval can be conducted. The retrieval performance for $k = \{1,10,20,50\}$ and all four model families can be found in figure \ref{fig:ridge}. The results share some characteristics with those presented in §\ref{sec:procrustes}, but in some cases performance drops for the families' largest models (e.g. OPT-6.7B and Ada-002), presumably because of poor signal-to-noise ratios in the extra dimensions, which, in Procrustes Analysis, are removed through principal component analysis. 

\vspace{-\baselineskip}
\vspace{-\baselineskip}
\subsection{Analogies}
\begin{figure}[h]
    \centering
    \includegraphics[width=13cm]{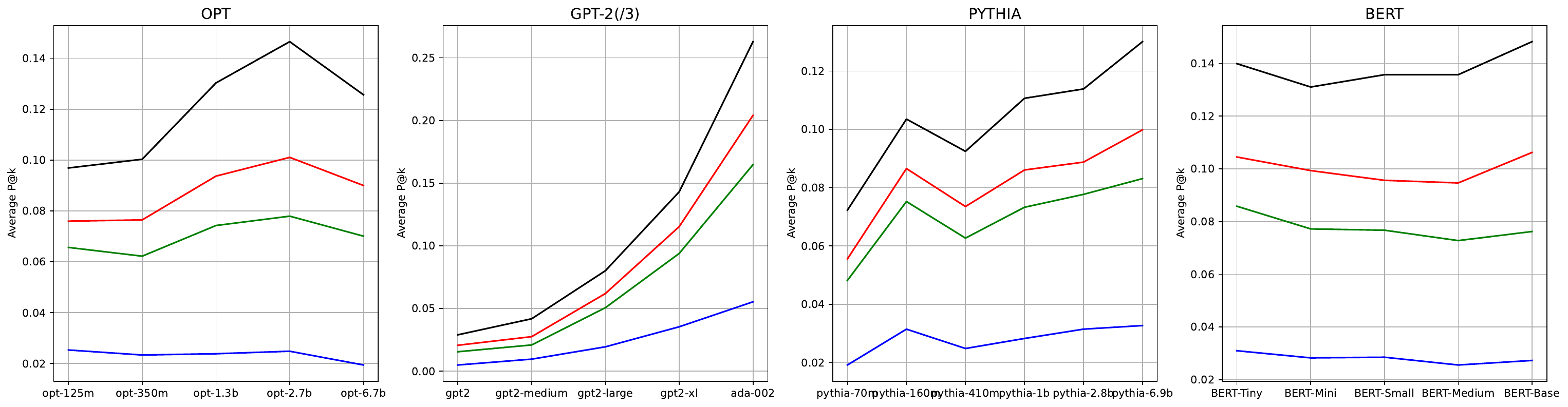}
    \caption{Plot labels: \textcolor{blue}{\rule[0.08cm]{0.5cm}{1.0pt} $k=1$}, \textcolor{green}{\rule[0.08cm]{0.5cm}{1.0pt} $k=10$}, \textcolor{red}{\rule[0.08cm]{0.5cm}{1.0pt} $k=20$}, \textcolor{black}{\rule[0.08cm]{0.5cm}{1.0pt} $k=50$}. The figure presents the results of how well the language models \textit{solve} the analogies from the WiQueen dataset and retrieve the correct the word. We report retrieval performance p@k at $k=\{1,10,20,50\}$ for 4 language model families.}
    \label{fig:analogy_res}
\end{figure}

Figure \ref{fig:analogy_res} shows the results of the four series of language models on the WiQueen data set. Again, we observe a consistent upward trend for all four model families. Note that this experiment more closely resembles a real-world task for a language model, as analogies play a central role in human commonsense reasoning \citep{ushio-etal-2021-bert}. The trends shown in figure \ref{fig:analogy_res} are in tune with those presented in §\ref{sec:procrustes}-\ref{sec:ref} and thus expands the evidence to support our arguments to more realistic use cases of language models.

\subsection{Representational Similarity Analysis}
The results of the representational similarity analysis for all four language model families can be found in the supplementary material. The partial convergence results are similar to those obtained with linear projection. 

\section{Analysis}

\subsection{LLMs and graph embeddings}
We first analyzed the robustness of our partial convergence results across various settings\footnote{96 total settings: 4 values of \textit{k}, 3 reference vector spaces, 4 LM families and 2 projection methods.}, fitting a linear trend line $y = mx + b$ to each convergence line using linear least squares regression. We saw that TransE and BigGraph had more convergence results than ComplEx: Whereas 0.969 of the convergence plots had positive slopes for TransE and BigGraph, only 0.875 of the convergence plots had positive slopes for ComplEx. We also saw that OPT and Pythia had the most robust convergence properties of the LLMs (with 1.0 of the convergence plots having positive slopes), compared to about 0.80 for GPT and 0.90 for BERT. Furthermore, we observe that when utilizing ridge regression as the projection method, the retrieval performance experiences a significant decline at the largest model in some cases. The largest models also correspond to those with the highest dimensional embeddings, which may suggest that this projection method encounters difficulties when the dimensionality of the input embeddings exceeds $d=2048$.

\subsection{Polysemy and semantic category}
We investigate the effect of polysemy and semantic category of the target words. Previous work on bilingual dictionary induction from cross-lingual vector space projections found high variance in retrieval scores across similar dimensions \cite{kementchedjhieva-etal-2019-lost,li2023implications}. Table \ref{tab:analysis_2} provides an overview of our results. Polysemy refers to the level of lexical ambiguity for target words, distinguishing between words with  one, two to three and four or more distinct meanings. We obtain polysemy counts for our target words (see \S\ref{sec:biggraph}) from NLTK's WordNet interface \citep{NLTK}. For semantic categories, we compare our experiments with common (frequent) words (see \S\ref{sec:biggraph}) to using only \textit{places} (geographic locations/places etc.) or \textit{names} (anthroponyms). The names were found in \footnote{\href{http://bitly.ws/DrMv}{US common names}} and the geographic locations were found in. \footnote{\href{http://bitly.ws/DrMJ}{World cities}}

\paragraph{Polysemy}
To investigate potential sources of error, we present statistics for three bins of polysemy counts in Table \ref{tab:analysis_2}. Our findings suggests that performance drops as the polysemy counts grows to more than four. This is intuitive because words with multiple meaning can be mapped to vastly different positions in the induced vector space, which might lead to a performance drop when retrieval is carried out in the reference vector space. These findings align well with those presented in \citep{li2023implications}, particularly that non-polysemous words tend to yield higher precision scores, but also that for some models (e.g. GPT-2), using words with two to three meanings results in better performance - these observations remain to be investigated in future work.
\paragraph{Semantic category}
Furthermore, we investigate the impact of semantic categories. We repeat the experiments across all settings using both the vocabulary of anthroponyms and world cities. The statistics presented in table \ref{tab:analysis_2} suggests that such semantic categories has an noticeable impact on performance. Specifically that anthroponyms has a relatively low max slope coefficient, suggesting slow convergence properties while the language model size grows. In addition to this, we observe that the \textit{places} (i.e. world cities) has a substantially higher positive rate, than other categories considered, which indicates that large language models might have better internal representations of some concepts compared to others.

\begin{table}[h]
\centering
\begin{tabular}{@{}llllllll@{}}
\toprule
\multicolumn{4}{l}{\textbf{Polysemy}}                                                                                                                                                                                                                                      & \multicolumn{4}{l}{\textbf{Semantic category}}                                                                                                                                                                                                                                \\ \midrule
\textbf{Variable}                                                          & \textbf{\% positive }                                                  & \textbf{Max coeff.}                                                    & \textbf{SD}                                                            & \textbf{Variable }                                                       & \textbf{\% positive}                                                   & \textbf{Max coeff. }                                                   & \textbf{SD}                                                            \\ \midrule
\begin{tabular}[c]{@{}l@{}}1\\ 2-3\\ 4+\end{tabular} & \begin{tabular}[c]{@{}l@{}}0.812\\ 0.792\\ 0.771\end{tabular} & \begin{tabular}[c]{@{}l@{}}0.172\\ 0.145\\ 0.129\end{tabular} & \begin{tabular}[c]{@{}l@{}}0.038\\ 0.032\\ 0.028\end{tabular} & \begin{tabular}[c]{@{}l@{}}Common\\ Places\\ Names\end{tabular} & \begin{tabular}[c]{@{}l@{}}0.938\\ 1.0\\ 0.917\end{tabular} & \begin{tabular}[c]{@{}l@{}}0.151\\ 0.105\\ 0.061\end{tabular} & \begin{tabular}[c]{@{}l@{}}0.029\\ 0.020\\ 0.013\end{tabular} \\ \bottomrule
\end{tabular}
\caption{The effect of polysemy and word classes on the convergence trend. \textit{Common} refer to common english words (i.e. those presented in §\ref{sec:biggraph}).}
\label{tab:analysis_2}
\end{table}

\begin{table}[h]
\centering
\begin{tabular}{@{}lllllllll@{}}
\toprule
\textbf{Models} & \textbf{Polysemy}                                    & \textbf{\begin{tabular}[c]{@{}l@{}}\tiny{BigGraph}\\ P@50\end{tabular}} & \multicolumn{1}{r}{\textbf{\begin{tabular}[c]{@{}r@{}}\tiny{TransE}\\ P@50\end{tabular}}} & \multicolumn{1}{r}{\textbf{\begin{tabular}[c]{@{}r@{}}\tiny{ComplEx}\\ P@50\end{tabular}}} & \textbf{\begin{tabular}[c]{@{}l@{}}Semantic\\ category\end{tabular}} & \multicolumn{1}{r}{\textbf{\begin{tabular}[c]{@{}r@{}}\tiny{BigGraph}\\ P@50\end{tabular}}} & \multicolumn{1}{r}{\textbf{\begin{tabular}[c]{@{}r@{}}\tiny{TransE}\\ P@50\end{tabular}}} & \multicolumn{1}{r}{\textbf{\begin{tabular}[c]{@{}r@{}}\tiny{ComplEx}\\ P@50\end{tabular}}} \\ \midrule
OPT-\small{6.7b}             & \begin{tabular}[c]{@{}l@{}}1\\ 2-3\\ 4+\end{tabular} & \begin{tabular}[c]{@{}l@{}}\textbf{0.595}\\ 0.490\\ 0.385\end{tabular}    & \begin{tabular}[c]{@{}l@{}}\textbf{0.748}\\ 0.680\\ 0.610\end{tabular}                       & \begin{tabular}[c]{@{}l@{}}\textbf{0.590}\\ 0.483\\ 0.435\end{tabular}                       & \begin{tabular}[c]{@{}l@{}}Common\\ Places\\ Names\end{tabular}      & \begin{tabular}[c]{@{}l@{}}0.203\\ \textbf{0.299}\\ 0.210\end{tabular}                        & \begin{tabular}[c]{@{}l@{}}\textbf{0.529}\\ 0.382\\ 0.264\end{tabular}                      & \begin{tabular}[c]{@{}l@{}}\textbf{0.444}\\ 0.325\\ 0.222\end{tabular}                       \\ \midrule
ADA-\small{002}                & \begin{tabular}[c]{@{}l@{}}1\\ 2-3\\ 4+\end{tabular} & \begin{tabular}[c]{@{}l@{}}\textbf{0.610}\\ 0.520\\ 0.423\end{tabular}    & \begin{tabular}[c]{@{}l@{}}\textbf{0.813}\\ 0.760\\ 0.675\end{tabular}                      & \begin{tabular}[c]{@{}l@{}}\textbf{0.618}\\ 0.528\\ 0.450\end{tabular}                       & \begin{tabular}[c]{@{}l@{}}Common\\ Places\\ Names\end{tabular}      & \begin{tabular}[c]{@{}l@{}}0.276\\ \textbf{0.373}\\ 0.285\end{tabular}                        & \begin{tabular}[c]{@{}l@{}}\textbf{0.651}\\ 0.465\\ 0.329\end{tabular}                      & \begin{tabular}[c]{@{}l@{}}\textbf{0.478}\\ 0.367\\ 0.250\end{tabular}                       \\ \midrule
Pythia-\small{6.9b}             & \begin{tabular}[c]{@{}l@{}}1\\ 2-3\\ 4+\end{tabular} & \begin{tabular}[c]{@{}l@{}}\textbf{0.495}\\ 0.373\\ 0.323\end{tabular}    & \begin{tabular}[c]{@{}l@{}}\textbf{0.538}\\ 0.488\\ 0.348\end{tabular}                      & \begin{tabular}[c]{@{}l@{}}\textbf{0.478}\\ 0.410\\ 0.310\end{tabular}                       & \begin{tabular}[c]{@{}l@{}}Common\\ Places\\ Names\end{tabular}      & \begin{tabular}[c]{@{}l@{}}0.126\\ \textbf{0.190}\\ 0.145\end{tabular}                        & \begin{tabular}[c]{@{}l@{}}\textbf{0.235}\\ 0.204\\ 0.150\end{tabular}                      & \begin{tabular}[c]{@{}l@{}}\textbf{0.228}\\ 0.188\\ 0.139\end{tabular}                       \\ \midrule
BERT-\small{BASE}               & \begin{tabular}[c]{@{}l@{}}1\\ 2-3\\ 4+\end{tabular} & \begin{tabular}[c]{@{}l@{}}\textbf{0.533}\\ 0.425\\ 0.320\end{tabular}    & \begin{tabular}[c]{@{}l@{}}\textbf{0.785}\\ 0.688\\ 0.538\end{tabular}                      & \begin{tabular}[c]{@{}l@{}}\textbf{0.578}\\ 0.505\\ 0.448\end{tabular}                       & \begin{tabular}[c]{@{}l@{}}Common\\ Places\\ Names\end{tabular}      & \begin{tabular}[c]{@{}l@{}}0.239\\ \textbf{0.350}\\ 0.259\end{tabular}                        & \begin{tabular}[c]{@{}l@{}}\textbf{0.508}\\ 0.359\\ 0.228\end{tabular}                      & \begin{tabular}[c]{@{}l@{}}\textbf{0.393}\\ 0.322\\ 0.198\end{tabular}                       \\ \bottomrule
\end{tabular}
\caption{Effect of polysemy and semantic category on the largest model from each model family. Procrustes is used as the projection method. Note that a low level of lexical ambiguity leads to better performance and that the best performing semantic category varies across the reference vector spaces.}
\label{tab:my-table}
\end{table}

\subsection{Analogies}
\citet{ushio-etal-2021-bert} investigated how well LLMs such as GPT are able to solve analogies. They obtained the best results using GPT. This aligns well with our finding that GPT-\{2,3\} has solid convergence properties and obtained the overall best results; see Figure \ref{fig:analogy_res}. 

\section{Discussion}
We have seen that language models converge on human-like concept organizations. How surprising is this? Given the contentious debate around whether large language models `understand' \cite{doi:10.1073/pnas.2215907120}, including whether they induce models of knowledge, our result is important. Large language models do not only learn to use patterns in context, but as a result, they induce compressed models of knowledge. In retrospect, it is also clear, however, that some results, e.g., the near-isomorphism of word vector spaces across languages \cite{vulic-etal-2020-good} or the near-isomorphism with representations from computer vision \cite{li2023implications}, already pointed in this direction. Language models for different languages likely learn similar concept geometries, because they induce models of (our knowledge of) the world. Language and computer vision models, in a similar way, share one reference, namely, the world we live in, and what we know about it.

\subsection{Practical implications}

There has already been considerable work on grounding language models in knowledge bases. This work often has focused on algorithms for joint language and graph embedding \cite{yu2022jaket}. Our work suggests that similar results can be obtained with `retro-fitting´ \cite{faruqui-etal-2015-retrofitting}, i.e., fine-tuning the language models to improve existing similarities. See, for example, the approach taken by \cite{Garneau2021AnalogyTM}. Our results also suggest, however, that in the limit, perhaps grounding in knowledge bases will become redundant. The systematicity of human-like conceptual organization in language models seemingly facilitates out-of-distribution capacities, e.g., enabling analogical inference. 

\subsection{Philosophical implications}

Our results clearly show that language models induce inferential semantics \cite{10.3389/frai.2021.682578,piantadosi2022meaning}. This, we believe, settles the debate about the capacities of large language models \cite{doi:10.1073/pnas.2215907120}. Our results also, however, question the divide between syntax and semantics \cite{Churchland1990CouldAM}. Semantics, in a way, seems to fall out of syntax. Clearly, our results have no bearing on intentionality (the aboutness of mental tokens), but they do suggest one way syntactic tokens acquire semantics. 

\section{Limitations}

We have experimented with three families of autoregressive language models and one non-autoregressive family. We have compared the word vector spaces induced by such models with three vector spaces induced by graph embedding algorithms over large knowledge bases. All our experiments have been limited to English. This, of course, is a major limitation. Language characteristics may effect the quality of word vector spaces, and morpho-syntactic properties may influence how corpora and knowledge bases align. Finally, while we do error analysis over polysemy and semantic categories, we acknowledge that the set of variables that covary with performance, is probably much larger. 

\section{Conclusion}

This paper weighs in on the debate around understanding in large language models and show how large language models converge toward human-like concept organization, building implicit models of the world (as we know it). Over 220 experiments, we show how language models converge toward human-like concept organization, with particularly strong similarities in how monosemous and common words are encoded. Our observations have important practical and philosophical implications, providing a possible explanation for the out-of-distribution capacities of large language models and settling the above debate. 
\section{Computational requirements}
A Google Colab Pro+ subscription or similar ($\geq52$GB RAM) is required in order to reproduce our experiments. We used an NVIDIA V100 and A100 Tensor Core GPU, provided by Google Colab.
\section{Acknowledgement}

\bibliography{custom}
\bibliographystyle{plainnat}
\end{document}